\documentclass{dialogue}
\pdfoutput=1

\begin{document}

\begin{otherlanguage}{english}
\begin{center}
{\Large\bfseries{LowResourceEval-2019: a shared task on morphological analysis for low-resource languages}}

\medskip

Klyachko E. L. (\texttt{elenaklyachko@gmail.com})

Institute of Linguistics, HSE, Moscow, Russia

\medskip
Sorokin A. A. (\texttt{alexey.sorokin@list.ru}) 

Moscow State University, Moscow, Russia

Moscow Institute of Physics and Technology, Dolgoprudny, Russia

\medskip

Krizhanovskaya N. B. (\texttt{nataly@krc.karelia.ru}) 

Krizhanovsky A. A. (\texttt{andrew.krizhanovsky@gmail.com}) 

Institute of Applied Mathematical Research of KarRC of RAS, Petrozavodsk, Russia
\medskip

Ryazanskaya G. M. (\texttt{galka1999@gmail.com}) 

Center for Language and Brain, HSE, Moscow, Russia

\end{center}

The paper describes the results of the first shared task on morphological analysis for the languages of Russia, namely, Evenki, Karelian, Selkup, and Veps. For the languages in question, only small-sized corpora are available. The tasks include morphological analysis, word form generation and morpheme segmentation. Four teams participated in the shared task. Most of them use machine-learning approaches, outperforming the existing rule-based ones. The article describes the datasets prepared for the shared tasks and contains analysis of the participants' solutions. Language corpora having different formats were transformed into CONLL-U format. The universal format makes the datasets comparable to other language corpura and facilitates using them in other NLP tasks.\medskip

\end{otherlanguage}

\bigskip

\begin{otherlanguage}{russian}
\begin{center}
{\Large\bfseries{LowResourceEval-2019: дорожка по морфологическому анализу для малоресурсных языков}}

\medskip

Клячко Е. Л. (\texttt{elenaklyachko@gmail.com})

Институт языкознания РАН, НИУ ВШЭ, Москва, Россия
\medskip

Сорокин А. А. (\texttt{alexey.sorokin@list.ru}) 

Московский Государственный Университет им. М.~В.~Ломоносова, Москва, Россия

Московский физико-технический институт, Долгопрудный, Россия

\medskip

Крижановская Н. Б. (\texttt{nataly@krc.karelia.ru}) 

Крижановский А. А. (\texttt{andrew.krizhanovsky@gmail.com}) 

Институт прикладных математических исследований КарНЦ РАН, Петрозаводск, Россия

\medskip
Рязанская Г. М. (\texttt{galka1999@gmail.com}) 

Центр языка и мозга, НИУ ВШЭ, Москва, Россия

\end{center}

В статье описывается первое соревнование, посвященное морфологическому анализу малоресурсных языков России, а именно: эвенкийского, карельского, селькупского и вепсского. Указанные языки располагают корпусами небольшого размера. Соревнование включало в себя автоматическое определение морфологических признаков, деление на морфемы, а также синтез словоформ. В статье описываются корпуса, специально подготовленные для соревнования, а также анализируются методы, использованные его участниками. Наилучшие результаты показали модели, основанные на нейронных сетях.

\medskip

\textbf{Ключевые слова:} морфологический анализ, морфемная сегментация, малые языки, малоресурсные языки
\end{otherlanguage}

\selectlanguage{english}
\section{Introduction}

According to the 2010 Census \cite{census}, more than 250 languages from 14 language families are spoken in Russia. About 100 of them are minority languages. It is worth noting that even non-minority languages, such as Yakut (Sakha), are considered vulnerable. For most languages of Russia, apart from Russian, digital resources either do not exist or are relatively scarce. 

The shared task\footnote{\url{https://lowresource-lang-eval.github.io/content/shared_tasks/morpho2019.html}} was held from January to March, 2019. The aims of the shared task were as following:

\begin{enumerate}
  \item to facilitate and stimulate the development of corpora and linguistic tools for minor languages:
  \begin{enumerate}
  \item One of the results produced by the shared task are text corpora which are uniformly tagged and accessible online.
  \item The participants are obliged to share the resulting systems.
  \end{enumerate}
  \item to inspire better communication between the communities of field linguists and NLP researchers;
  \item to figure out how modern methods of morphological analysis, tagging, segmentation, and synthesis cope with sparse training data, the lack of standard language and large rate of dialectal varieties.
\end{enumerate}

\section{Related work}

We present a short survey of corpora for minor languages of Russia. A detailed survey of Russian minority language corpora and morphology tools as of 2016 can be found in \cite{Arkhangelskiy2016DevelopingMA}. However, more corpora have been developed since then. Therefore we suppose that the topic should be revisited.

\subsection{Corpora}
Most corpus resources are created by language activists and are based on digitalized books and other printed materials. Some examples are the corpora created by The Finno-Ugric Laboratory for Support of the Electronic Representation of Regional Languages\footnote{\url{http://fu-lab.ru/}}, The digital portal of Selkup language\footnote{\url{http://selkup.org/}} etc.

On the other hand, field data collected during linguistic expeditions is often transformed into corpora. These corpora are usually created by universities such as the corpora published at HSE Linghub\footnote{\url{https://linghub.ru/}}, VepKar\footnote{\url{http://dictorpus.krc.karelia.ru/en}} at the Karelian Research Center, the Siberian-Lang language data\footnote{\url{http://siberian-lang.srcc.msu.ru/}} collected by MSU and The Institute of Linguistics (Russian Academy of Sciences) and many other projects. Furthermore, some projects also leverage old field data, digitizing it. For example, in INEL project\footnote{\url{https://inel.corpora.uni-hamburg.de/}}, field data for Selkup and now extinct Kamassian language have been digitized and processed.

Field data consists mostly of oral texts. Therefore, corpus materials often contain non-standard varieties of a language and demonstrate remarkable dialectal and sociolinguistic features. However, the high level of variation makes it challenging for automatic processing.

Table \ref{tab:resource} shows the resources for the languages of Russia, which are available online as language corpora with search facilities. We do not include published books of interlinearized texts although they can be used as a source for future corpora.

\begin{longtable}[ht!]{|p{2cm}|p{3cm}|p{2cm}|l|p{2cm}|}
\caption{Morphological resources for languages of Russia}\label{tab:resource}
\\\hline
\textbf{Language} & \textbf{Tokens} & \textbf{Parallel languages} & \textbf{Markup} & \textbf{License} \\ \hline
Abaza\footnote{https://linghub.ru/abaza\_rus\_corpus/search} & 32 796 & Russian & no tags & NA \\ \hline
Avar\footnote{http://web-corpora.net/AvarCorpus/search/?interface\_language=en} & 2 300 000 & - & tags, no disambiguation & NA \\ \hline
Adyghe\footnote{http://adyghe.web-corpora.net} & 7 760 000 & Russian & tags, no disambiguation & NA \\ \hline
Archi\footnote{http://web-corpora.net/ArchiCorpus/search/index.php?interface\_language=en} & 58 816 & - & tags, not detailed & NA \\ \hline
Bagvalal\footnote{http://web-corpora.net/BagvalalCorpus/search/?interface\_language=en} & 5 819 & Russian & tags with disambiguation & NA \\ \hline
Bashkir\footnote{http://bashcorpus.ru/bashcorpus/} & 20 584 199 & - & tags, no disambiguation & NA \\ \hline
Beserman Udmurt\footnote{ http://beserman.ru/corpus/search} & 65 000 & Russian & tags with disambiguation & CC BY 4.0 \\ \hline
Buryat \footnote{http://web-corpora.net/BuryatCorpus/search/?interface\_language=en} & 2 200 000 & - & tags, no disambiguation & NA \\ \hline
Chukchi\footnote{ http://chuklang.ru/corpus} & 6393 & English, Russian & tags with disambiguation & NA \\ \hline
Chuvash \footnote{http://corpus.chv.su} & 1 147 215 & Russian (partially) & no tags & NA \\ \hline
Crimean Tatar\footnote{ https://korpus.sk/QIRIM} & 56 752 & - & no tags &  \\ \hline
Dargwa \footnote{http://web-corpora.net/SanzhiDargwaCorpus/search/?interface\_language=en} & 48 957 & Russian & tags with disambiguation & NA \\ \hline
Erzya \footnote{http://erzya.web-corpora.net} &  3 130 000 & partially (Russian) & tags, no disambiguation & CC BY 4.0 \\ \hline
Evenki \footnote{http://corpora.iea.ras.ru/corpora/news.php?tag=6} & 121 286 & Russian (partially) & tags, no disambiguation & own license \\ \hline
Evenki \footnote{http://gisly.net/corpus} & 25 000 & Russian & tags with disambiguation & NA \\ \hline
Godoberi\footnote{ http://web-corpora.net/GodoberiCorpus/search/?interface\_language=en} & 872 & English & tags with disambiguation & NA \\ \hline
Kalmyk\footnote{http://web-corpora.net/KalmykCorpus/search/?interface\_language=en} & 858 235 & - & tags, no disambiguation & NA \\ \hline
Karelian\footnote{ http://dictorpus.krc.karelia.ru/en} & 66 350 & Russian & tags with disambiguation (partial) & CC BY 4.0\\ \hline
Khakas\footnote{http://khakas.altaica.ru} & 285 000 & Russian & tags, no disambiguation & NA \\ \hline
Khanty\footnote{https://kitwiki.csc.fi/twiki/bin/view/FinCLARIN/KielipankkiAineistotKhantyUHLCS} & 161 224 & Finnish, English, Russian & tags with disambiguation (partial) & CLARIN RES \\ \hline
Komi-Zyrian\footnote{ http://komicorpora.ru} & 54 076 811 & - & tags, no disambiguation & NA \\ \hline
Mansi\footnote{http://digital-mansi.com/corpus} & 961 936 & - & tags (not detailed) & NA \\ \hline
Nenets \footnote{http://corpora.iea.ras.ru/corpora} & 125 421 & Russian (partially) & no tags & own license \\ \hline
Nenets \footnote{http://www.ling.helsinki.fi/uhlcs/metadata/corpus-metadata/uralic-lgs/samoyedic-lgs/nenets} & 148 348 & Russian & no tags & CLARIN RES \\ \hline
Ossetic \footnote{ http://corpus.ossetic-studies.org/search/index.php?interface\_language=en} & 12 000 000 & - & tags, no disambiguation & NA \\ \hline
Romani \footnote{http://web-corpora.net/RomaniCorpus/search/?interface\_language=en} & 720 000 & - & tags, no disambiguation & NA \\ \hline
Selkup \footnote{https://corpora.uni-hamburg.de/hzsk/en/islandora/object/spoken-corpus\%3Aselkup-0.1} & 18 763 & English, German, Russian & tags with disambiguation & CC BY-NC-SA 4.0 \\ \hline
Shor \footnote{http://corpora.iea.ras.ru/corpora/news.php?tag=3\&amp;period=} & 262 153 & Russian (partially) & 1 & own license \\ \hline
Tatar \footnote{http://tugantel.tatar} & 180 000 000 & - & tags, no disambiguation & NA \\ \hline
Udmurt \footnote{http://web-corpora.net/UdmurtCorpus/search/?interface\_language=en} & 7 300 000 & - & tags, no disambiguation & NA \\ \hline
Veps \footnote{http://dictorpus.krc.karelia.ru/en} & 46 666 & Russian & tags with disambiguation (partial) & CC BY 4.0 \\ \hline
Yiddish \footnote{http://web-corpora.net/YNC/search} & 4 895 707 & - & tags, no disambiguation & NA \\ \hline

\end{longtable}

\subsection{Other shared tasks on low-resource evaluation}

The main Shared Task concerned with morphological tagging is the well-known CoNLL Shared Task on parsing from raw data to Universal Dependencies \cite{zeman2017conll}\footnote{http://universaldependencies.org/conll17/}, \cite{zeman2018conll}\footnote{http://universaldependencies.org/conll18/}. Though the main goal of this competition is evaluation of dependency parsers, it also deals with morphological analysis since morphological tags are used as features for further syntactic processing. The task included 82 corpora of different size for 57 languages in its 2017 edition, with the size of the corpora ranging from several hundred words to more than 1 mln. The shared task organizers name $9$ treebanks as small (they contain from $4$K to $20$K words) and $9$ as low-resource. The size of low-resource treebanks is less than $1000$ words. These two categories of treebanks differ dramatically in terms of tagger performance: while the average accuracy of morphological tagging was $82\%$ for small treebanks, the quality for low-resource language was only $25\%$. Therefore, the datasets used in our Shared Task better fit to ``small'' category than to the low-resource one. However, for most small treebanks in UD one can also learn from data either for a closely related language, for example, Finnish for North S\'{a}mi (sme\_giella), or even a different corpus for the same language (Latin la\_proiel corpus with 272K words for la\_perseus corpus with 18K words). That is not the case for two main languages of our Shared Task, Evenki and Selkup, while for Veps and Karelian one can use Finnish or Estonian as an additional source.

The only known competition on morpheme segmentation was MorphoChallenge Shared Task \footnote{\url{http://morpho.aalto.fi/events/morphochallenge/}} held from $2005$ to $2010$. The amount of labeled training data in its edition was rather small (about 1700 word types), however, the organizers provided an additional word list, which included several hundred thousands of unsegmented words since morpheme segmentation was usually treated as minimally supervised or semi-supervised problem. In recent studies on supervised morpheme segmentation, such as \cite{kann2018fortification}, the training dataset usually did not exceed $2000$ words, though the amount of segmented data in our competition was even greater than in analogous studies.

The main Shared Task on morphological inflection, the Sigmorphon Shared Task \cite{cotterell2018conll} specially provided three types of training datasets: low-resource (100 words), middle (1000 words) and large (up to 10000 words). 

\section{Shared task description}

The task consists of three tracks which are described below. Evaluation scripts can be found in our Github repository\footnote{\url{https://github.com/lowresource-lang-eval/morphology_scripts/tree/master/evaluation}}.
The participants were allowed to use any external dataset. However, they were required to publish their solutions into open-source. It was done to accelerate NLP tool development for minor languages. The participants could provide several solutions.

\subsection{Morphological analysis}

The morphological analysis task was to to produce lemmata, part-of-speech tags and morphological features for tokenized sentences. Training data was annotated using CONLL-U format\footnote{https://universaldependencies.org/format.html} also used in Universal Dependencies project. We extended the annotation for the corpora without morphological disambiguation: in case a word had several analyses in corpus, we listed them all on consecutive lines. Words lacking morphological analysis in the corpora were annotated with distinguished \textsc{unkn} tag. An example of the markup for Karelian can be found below:
\medskip

13	julkaistuja	UNKN	\_	\_	\_	\_	\_	\_	\_

14	kirjoja	kirja	NOUN	\_	Number=Plur|Case=Par	\_	\_	\_	\_

\medskip
The following metrics were evaluated:
\begin{enumerate}
    \item the fraction of word forms with correct lemmata;
    \item the fraction of sentences where all word forms have correct lemmata;
    \item the fraction of word forms with correct part-of-speech tags;
    \item the fraction of sentences where all word forms have correct part-of-speech tags;
    \item precision, recall, and F1 score of predicted morphological features calculated according to:
 
 $$
 \begin{array}{rcccl}
 P & = & \dfrac{TP}{TP+FP}, \\[16pt]
 R & = & \dfrac{TP}{TP+FN}, \\[16pt]
 F1 & = & \dfrac{2 PR}{P + R} & = & \dfrac{TP}{TP + 0.5(FP+FN)},
 \end{array}
 $$
   
where \textit{TP} is the number of true positives (correct morphological features),  \textit{FP} is the number of false positives (incorrectly assigned morphological features) and \textit{FN} is the number of false negatives (missed morphological features).

\end{enumerate}

\subsection{Morpheme segmentation}

The training data consisted of tokenized sentences with each word split into morphemes. The task was to train a model which could produce morpheme segmentation for unknown words, too. Model quality was evaluated similarly to MorphoChallenge\footnote{\url{http://morpho.aalto.fi/events/morphochallenge2005/evaluation.shtml}}, i. e. we calculated boundary precision (\textit{P}), recall(\textit{R}), and F1 score using traditional formulas, where true positives are correct boundaries,  false positives are incorrectly predicted boundaries and false negatives are missed boundaries.

\subsection{Morpheme synthesis}
The training data was the same as in the morphological analysis task. The participants were to generate word forms, given lemmata, part-of-speech tags, and other morphological tags. The following measures were calculated:
\begin{enumerate}
    \item the fraction of word forms which were absolutely correct;
    \item average Levenshtein distance between the word forms generated by the participant and the correct word forms. If several are possible, the closest one is used.
\end{enumerate}

\section{Evaluation datasets}
The following datasets were kindly provided by their creators:
\begin{enumerate}
    \item Evenki: mainly oral texts recorded in 1998—2016 during fieldwork trips by Olga Kazakevich et al. Has morphological information as well as morpheme segmentation \cite{kazakevich2013sozdanie4670818};
    \item Selkup: oral texts recorded by A. I. Kuzmina in 1962—1977, processed and annotated within the INEL project. Has morphological information as well as morpheme segmentation \cite{orlova2018inel164881580};
    \item Veps and Karelian corpus developed within the VepKar project (described in more detail below).
\end{enumerate}
It is worth noting that the corpora have been created by linguists and are based on fieldwork data with detailed manual markup. This is unusual for ordinary computational linguistics corpora for major languages, which are usually based on written sources and are therefore more standardized and balanced.

All the corpora had different formats and incompatible markup standards. This would make it hard for the participants to use them. Furthermore, our aim was to allow the participants to combine data from different corpora, using transfer learning or other methods. Therefore, the corpora were converted into the morphological CONLL-U format, used in the Universal Dependencies (UD) treebanks \footnote{\url{https://universaldependencies.org/}}. Moreover, using the standard format makes the resources of the languages in question accessible to researchers all over the world. It makes it possible to include them into the community of "living" and "major" languages (such as Russian and English), which are available to researchers all over the world for processing and building computational models. 

On the one hand, the morphological annotation of UD is quite scarce due to the principles of its construction (new tags are only added after the treebanks are added to the project). As a result, we had to exclude many morphological tags. In some linguists' opinion, the resulting narrowed format deprived the language data of essential linguistic information. However, we regarded the narrowing as a necessary trade-off. In addition, the necessity to reformat the corpora made us reanalyze some complex cases and find mistakes in the analysis.

The complex export process is described below in greater detail. 
\subsection{Export of the Evenki corpus to CONLL format}
The Evenki corpus data consisted of EAF\footnote{\url{https://tla.mpi.nl/tla-news/documentation-of-eaf-elan-annotation-format/}} format files. Some texts used in the corpora were originally manually annotated interlinearized texts, lacking lemmata and POS tags. Determining those was the most difficult part of the corpus transformation process, and involved manual work.
For instance, the Evenki corpus contained word forms like \textit{oldomotto:wer} (`in order to catch fish`), with the derivative suffix \textit{-mo}(`hunt`)  attached to the nominal \textit{oldo} (`fish`) stem. When preparing the data, we turned this combination of morphemes into a single lemma, namely \textit{oldomo}.

\subsection{Export of the Selkup corpus to CONLL format}
In contrast with the Evenki corpus, the INEL Selkup corpus contained the necessary data. The difficulty of the transformation process consisted in the mapping between the rich and detailed corpus markup and the CONLL format. It was also troublesome to determine the lemma. Our criterion for lemma determination was to combine the stem and the derivative affixes but not the inflectional ones. Thanks to the help of experts, we could distinguish between the two sets of affixes. For example, we considered some aspectual affixes to be inflectional for Evenki, according to the grammars. In contrast with it, Selkup aspectual morphemes were considered to be derivative. For example, \textit{kurol`na} was segmented as \textit{kur-ol`(INCH)-na(CO.3SG.S)}. Therefore, the first two morphemes were considered to constitute the lemma \textit{kurol`}.

\subsection{Export of the VepKar corpus to CONLL format}
\subsubsection{Languages and dialects of the VepKar corpus}
For the shared task, the VepKar developers have presented texts in Veps language and in three main supradialects of Karelian language. 

\begin{figure}[htbp]
\centering
     \includegraphics[width=1.0\textwidth]{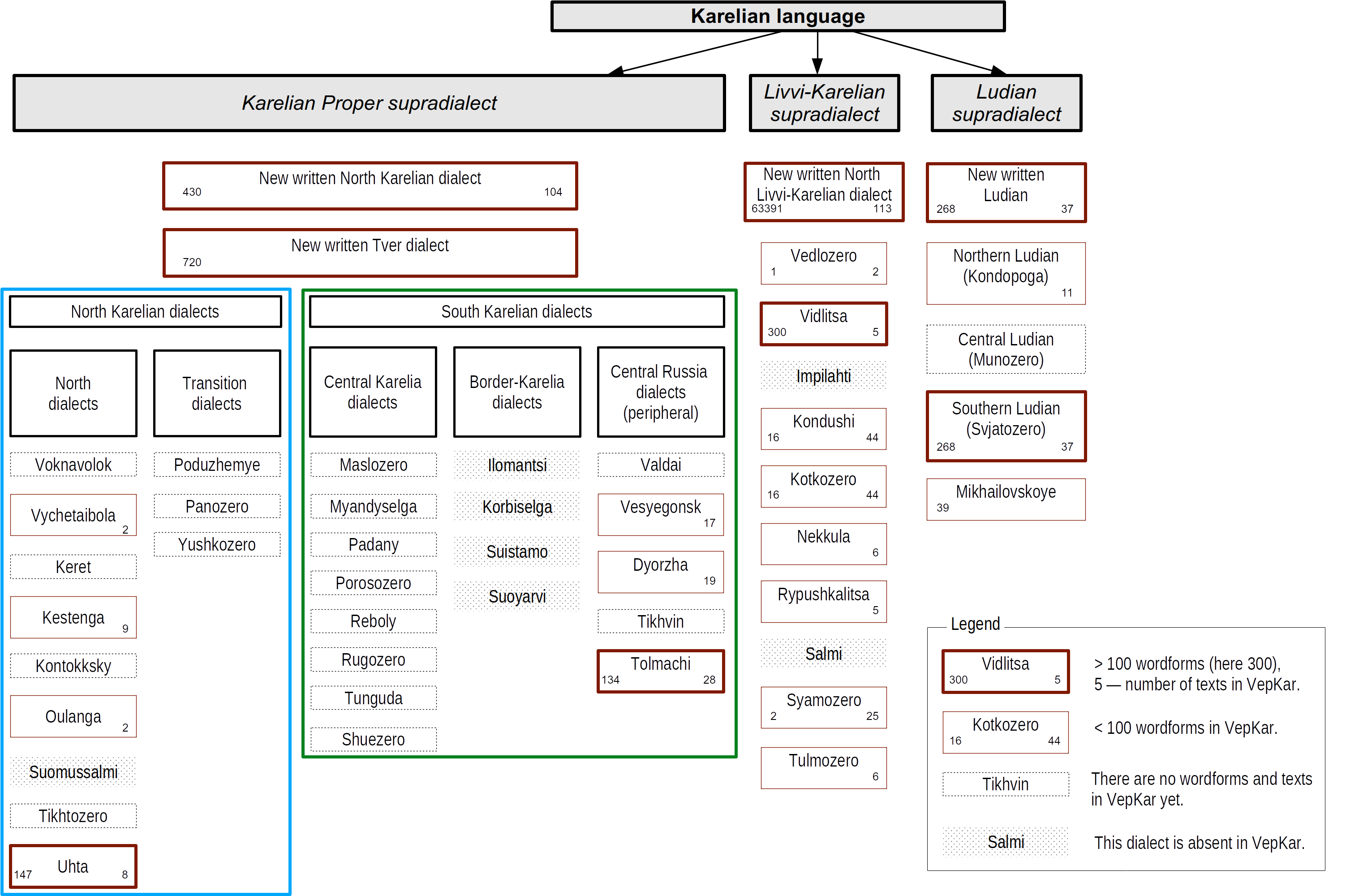}
\caption{Scheme of dialects of the Karelian language, the number of wordforms (left) and the number of texts (right) in these dialects in the VepKar corpus
\label{fig:karelian_dialects}
}
\end{figure}

VepKar contains a variety of dialects and subdialects of the Karelian language (see fig. \ref{fig:karelian_dialects}). 
The scheme is based on \cite{zaykov2000glagol}, \cite{novak2019}.

There are three written Karelian standard languages. This is due to several reasons. Native Karelian speakers live on a rather vast territory. For several centuries, the language has been influenced by the neighboring Veps, Finnish, and, of course, Russian. The lexical and phonetic systems were the ones most influenced from the outside. This influence gave rise to the three supradialects. Therefore, the corpus uses a separate Karelian dictionary for each supradialect. As of February, 2019 the statistics for the corpus were as following:
\begin{enumerate}
    \item Olonets Karelian or Livvi (17 thousand lemmata);
    \item Ludic Karelian (500 lemmata); 
    \item Karelian Proper (100 lemmata).
\end{enumerate}

Therefore, three export data sets in CONLL format have been generated, one for each dialect.

\subsubsection{The most frequent tokens list}
\begin{figure}[htbp]
\centering
     \includegraphics[width=0.8\textwidth]{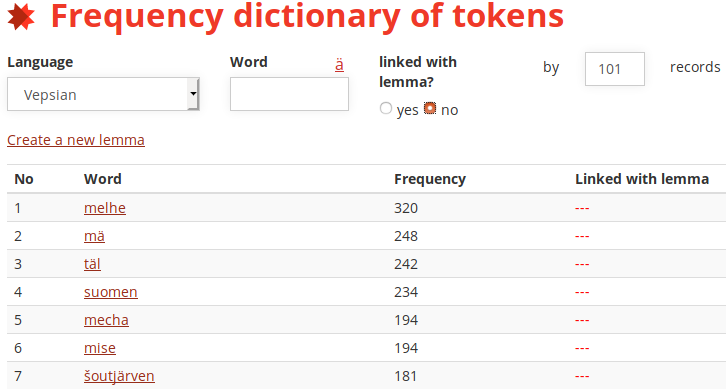}
\caption{The first most frequent tokens of Veps texts in the VepKar corpus which are absent in the VepKar dictionary, with links to usage examples and frequencies in the corpus}
\label{fig:vepkar_frequency}
\end{figure}

To assist lexicographers in their manual markup process, a generator of the most frequent tokens list (words from the corpus texts) was developed. It is available at the VepKar website\footnote{\url{http://dictorpus.krc.karelia.ru/ru/corpus/word/freq_dict}}. 
Using the radio button “does this word exist in the dictionary?”, one can get a list of the most frequent tokens that do not have dictionary entries in the VepKar dictionary (see fig. \ref{fig:vepkar_frequency}).

This list allows the corpus editors to add the most frequent word forms to the corpus dictionary first. Processing primarily the most frequent word forms accelerates the morphological markup of most corpus texts.
In the dictionary, two options are possible:
\begin{itemize}
    \item There is a lemma and there is no word form.
    \item There is neither word form nor lemma.
\end{itemize}

In the second case, while creating the lemma, the editor can also add the other word forms. This makes text markup possible if the lemma is found in texts in other grammatical forms.

\subsubsection{VepKar development}
Participation in this competition was the driving force for the development of the VepKar corpus. To export the data from the VepKar corpus to CONLL, the corpus structure had to be refined significantly. The following features were added to the morphological properties of a lemma:
\begin{itemize}
    \item for nouns: animacy (“Animacy” in Universal features);
 \item pluralia tantum (Number=Plur);
\item for verbs: transitivity (Subcat);
\item for numerals: type of numeral (NumType: quantitative, collective, ordinal, fractional);
\item for pronouns:  type of pronoun (PronType); 
\item for adjectives and adverbs: degree of comparison (Degree);
\item for adverbs: type of adverb (AdvType).
\end{itemize}
Initially, VepKar had one part of speech to designate conjunctions. According to the Universal POS tags, the VepKar conjunctions were divided into subordinating and coordinating.
\subsubsection{The exporting process}

While exporting VepKar data to CONLL format, the following conventions were accepted:
\begin{enumerate}
    \item For an unknown lemma, write UNKN to the LEMMA column and an underscore to the remaining columns.
    \item Write each pair of LEMMA + UPOS on separate lines.
    \item Export prepositions (PREP) and postpositions (POSTP) from VepKar corpus to ADP in CONLL. Features PREP or POSTP are indicated in the XPOS field.
    \item In a multilingual corpus one file is generated for one language.
    \item CONLL-style comments are used for adding sentence identifiers.
\end{enumerate}
\subsubsection{Corpus data not included in the CONLL export}

In the VepKar corpus there are data that were not exported to CONLL: predicatives (23 lemmas in Olonets Karelian) and phraseological units. 

\clearpage
\subsection{Dataset statistics}
Table \ref{tab:statistics} summarizes some statistical features of the datasets:

\begin{longtable}{lp{2cm}rrrrp{1.2cm}p{1.5cm}p{1cm}p{1cm}}
\caption{Dataset statistics}
\\\hline Part & Language &  Sentences & Words & POSes & Tags\footnote{Pairs of feature=value} & Rare(3) tags\footnote{occurring less than 3 times in the train set} & Rare(10) tags\footnote{occurring less than 10 times in the train set} & Full  tags\footnote{combinations of tags} & Rare\footnote{less than 10\% of all words}
full tags \% \\\hline
Train & Evenki & 5 527 & 26 926 & 12 & 55 & 0 & 2 & 714 & 100 \\
Test & Evenki & 548 & 2 819 & 12 & 53 & 0 & 1 & 270 & 98 \\
Train & Selkup & 2 394 & 13 436 & 12 & 34 & 2 & 3 & 218 & 98 \\
Test & Selkup & 425 & 2 426 & 12 & 30 & 1  & 1 & 109 & 95 \\
Train & Veps & 38 793 & 357 811 & 13 & 47 & 0 & 4 & 147 & 99 \\
Test & Veps & 2 163 & 19 376 & 13 & 42 & 0 & 1 & 86 & 100 \\
Train & Karelian (proper) & 7 048 & 68 296 & 11 & 34 & 4 & 6 & 66 & 100 \\
Test & Karelian (proper) & 919 & 8 640 & 9 & 30 & 1 & 3 & 44 & 100 \\
Train & Karelian (Ludic) & 1 711 & 15 805 & 12 & 27 & 5 & 5 & 29 & 97 \\
Test & Karelian (Ludic) & 204 & 1 968 & 11 & 22 & 0 & 0 & 19 & 95 \\
Train & Karelian (Livvi) & 6 213 & 57 093 & 13 & 23 & 4 & 6 & 26 & 88 \\
Test & Karelian (Livvi) & 745 & 7 206 & 13 & 17 & 0 & 1 & 19 & 84
\label{tab:statistics}
\end{longtable}91.2 
One can see from this table that the concept of "full tags", i.e. sets of morphological features, does not seem to work well for agglutinating languages due to the huge number of combinations.

\section{Participants and results}

\subsection{System description.}

In the morphological analysis track, three teams took part. 
The morpheme segmentation and word form generation tasks were less popular with only one team participating in each of them.

\textbf{MSU-DeepPavlov} team \cite{sorokin2019} utilized recurrent neural networks on word level. The embeddings of words were obtained using convolutional networks with highway layer on the top, closely following \cite{heigold2017extensive} and \cite{sorokin2018improving}. This team demonstrated the highest scores on Evenki and Selkup datasets on all metrics and on Veps dataset on part-of-speech prediction\footnote{It did not participate on other subtasks.}.

The second team, \textbf{drovoseq} used BertBiLSTMAttnNMT encoder-decoder architecture to decode the optimal sequence of morphological tags. The third one, \textbf{SPBUMorph} used a variant of Markov models to evaluate the probability of a tag given the word. 

For lemmatization, the winning \textbf{MSU-DeepPavlov} team used a neural network to predict the pattern of the transformation between the surface word and its initial form, while \textbf{drovoseq} used encoder-decoder architecture.

The only submitted morpheme segmentor used the model similar to \cite{sorokin2018deep}, which reduced the task of morpheme segmentation to sequence labeling.

\subsection{Results}

The results are shown in Tables \ref{Morphological analysis: results}, \ref{tab:morpheme} and \ref{tab:generation}:

\begin{longtable}[ht!]{p{2cm}p{1cm}p{1.5cm}p{1cm}p{1cm}p{1cm}p{1cm}p{1cm}p{1cm}p{1cm}}
\caption{Morphological analysis: results}
\label{Morphological analysis: results}
\\\hline Team & \# & Language & \% of correct lemmata for wf & \% of correct lemmata for sentences & \% of correct POS'es for wf & \% of correct POS'es for sentences & feature precision & feature recall & feature F2 \\\hline
drovoseq & 1 & Evenki & 0,8617 & 0,7550 & 0,8811 & 0,8086 & 0,8112 & 0,7993 & 0,8052 \\
drovoseq & 1 & Karelian (proper) & 0,9971 & 0,9869 & 0,9909 & 0,9603 & \textbf{0,9539} & \textbf{0,9373} & \textbf{0,9455} \\
drovoseq & 1 & Karelian (Ludic) & 0,9959 & 0,9828 & 0,9726 & 0,8897 & \textbf{0,9356} & \textbf{0,8769} & \textbf{0,9053} \\
drovoseq & 1 & Karelian (Livvi) & 0,9629 & 0,8631 & 0,8969 & 0,7168 & \textbf{0,8471} & \textbf{0,7985} & \textbf{0,8221} \\
drovoseq & 1 & Selkup & 0,8780 & 0,7647 & 0,8343 & 0,7529 & 0,8014 & 0,7713 & 0,7861 \\
drovoseq & 1 & Veps & 0,9761 & 0,9087 & 0,9572 & 0,8534 & \textbf{0,6691} & 0,4666 & 0,5498 \\
drovoseq & 2 & Evenki & 0,8710 & 0,7620 & 0,9075 & 0,8201 & 0,8156 & 0,8217 & 0,8187 \\
drovoseq & 2 & Karelian (proper) & 0,9977 & 0,9889 & 0,9992 & \textbf{0,9956} & 0,4045 & 0,1776 & 0,2468 \\
drovoseq & 2 & Karelian (Ludic) & 0,9970 & \textbf{0,9851} & \textbf{0,9959} & \textbf{0,9777} & 0,5962 & 0,3570 & 0,4466 \\
drovoseq & 2 & Karelian (Livvi) & \textbf{0,9797} & 0,9108 & \textbf{0,9781} & \textbf{0,9074} & 0,6751 & 0,4489 & 0,5392 \\
drovoseq & 2 & Selkup & 0,8941 & 0,7759 & 0,8586 & 0,7354 & 0,8029 & 0,8026 & 0,8028 \\
drovoseq & 2 & Veps & \textbf{0,9875} & \textbf{0,9480} & 0,9938 & 0,9753 & 0,4873 & 0,2678 & 0,3457 \\
MSU-DeepPavlov & 1 & Evenki & \textbf{0,8838} & \textbf{0,7914} & \textbf{0,9122} & \textbf{0,8421} & \textbf{0,8805} & \textbf{0,8809} & \textbf{0,8807} \\
MSU-DeepPavlov & 1 & Selkup & \textbf{0,9031} & \textbf{0,8035} & \textbf{0,8957} & \textbf{0,7965} & \textbf{0,9095} & \textbf{0,9082} & \textbf{0,9089} \\
MSU-DeepPavlov & 1 & Veps & 0,3003 & 0,5146 & \textbf{0,9943} & \textbf{0,9769} & 0,5471 & \textbf{0,8073} & \textbf{0,6522} \\
SPBUMorph & 1 & Evenki & 0,7125 & 0,2857 & 0,7222 & 0,3099 & 0,1503 & 0,3692 & 0,2137 \\
SPBUMorph & 1 & Karelian (proper) & \textbf{0,9992} & \textbf{0,9913} & \textbf{0,9994} & 0,9935 & 0,7028 & 0,9172 & 0,7958 \\
SPBUMorph & 1 & Karelian (Ludic) & 0,9975 & 0,9706 & 0,9959 & 0,9608 & 0,5692 & 0,8674 & 0,6873 \\
SPBUMorph & 1 & Karelian (Livvi) & 0,9653 & 0,7369 & 0,9460 & 0,6148 & 0,4742 & 0,7064 & 0,5674 \\
SPBUMorph & 1 & Selkup & 0,6834 & 0,2000 & 0,6818 & 0,2447 & 0,1400 & 0,3147 & 0,1938 \\
SPBUMorph & 1 & Veps & 0,9839 & 0,8798 & 0,9899 & 0,9177 & 0,5471 & 0,8073 & 0,6522 \\
\end{longtable}

\begin{longtable}[ht!]{lllllll}
\caption{Morpheme segmentation: results}\label{tab:morpheme}
\\\hline Team & \# & Language & Precision & Recall & F1 & \% of totally correct wordforms \\\hline
deeppavlov & 1 & Evenki & 0,9774 & 0,9783 & 0,9779 & 0,9317 \\
deeppavlov & 1 & Selkup & 0,9538 & 0,9551 & 0,9544 & 0,8640
\end{longtable}

\begin{longtable}[ht!]{lllll}
\caption{Word form generation: results}\label{tab:generation}
\\\hline
Team & \# & Language & Totally correct & Averaged Levenshtein distance \\\hline
SAG\_TEAM & 1 & Evenki & 0,5325 & 1,2585 \\
SAG\_TEAM & 1 & Selkup & 0,5076 & 1,1621
\end{longtable}

\section{Discussion}

For the time being, the languages under consideration do not have robust rule-based parsers, therefore the only source  of comparison is the annotation of the test set. We also notice that participants reported errors and discrepancies in the annotation during training phase. Although we fixed most of them after the discussion, this could potentially influence the systems' efficiency.

First, we would like to note that the topmost system achieved significantly high scores on tasks of morphological analysis, lemmatization and morpheme segmentation, which is comparable to scores of state-of-the-art systems on other datasets of similar size. 

\subsection{Morphological analysis results}

It is interesting that the systems made similar mistakes. It certainly has to do with the limitations of the data itself, its relatively low amount and scarcity. However, the percentage of errors differs significantly between the systems, which implies that different models require different amounts of labeled data to be trained on.

\subsubsection{Lemmatization}
Lemmatization errors can be grouped as following:
\begin{enumerate}
    \item Rare lemmata: e. g., the Evenki \textit{jaja} `to chant shamanic songs` can only be found in few texts.
    \item "Non-standard" lemmata: the oral texts in a minor language naturally contain a lot of loanwords. These loanwords, especially recent ones, are often different phonetically from the basic words. They seem to present troubles for all systems. E. g., the Evenki \textit{kirest} `cross` < Russian \textit{krest} has \textit{st} consonant cluster, which is not typical for an Evenki word.
    Another example is \textit{penśianerka} 'pensionnaire (woman)' < Russian \textit{penśianerka}. This word with its inital "p" sound is not typical for the language. Furthermore, its ending corresponds with the \textit{-rkV} suffix. Not surprisingly, most systems judged \textit{-rka} to be a suffix in this word.
    Similarly, the systems split the Selkup word \textit{poshalusta} 'please' < Russian \textit{pozhalusta}, separating the ending \textit{sta}.
    It would be interesting to check if the results could improve if the systems accounted for Russian loanwords.
    \item Short lemmata. The systems seemed to prefer long roots over short ones. As a result, word forms with one-letter roots are processed incorrectly. For example, \textit{e} 'negative verb' or \textit{i:} 'enter' presented a trouble for the systems.
    On the other hand, in some cases, the lemmata were standard and quite wide-spread. However, their ending in a letter which itself constituted a wide-spread suffix caused the systems to incorrectly split the lemma. In the Evenki data, this is true for \textit{l}, \textit{n} or \textit{t}.
    \item Morphophonological phenomena were difficult to follow for the systems. E. g., in \textit{uguchak-ker} 'reindeer-RFL.PL' the \textit{-ker} part is a surface realization of the \textit{-wer} morpheme after \textit{k}. Similar \textit{kw} -> \textit{kk} alternations can be found in the training data. However, the systems could not grasp this alternation.
\end{enumerate}

\subsubsection{Determining POS}

As regards the POS determination, the errors show that the systems could not reliably distinguish between nominal and verbal categories. One could expect the systems to confuse \textit{nouns} and \textit{adjectives} but not \textit{nouns} and \textit{verbs}. However, it is the \textit{VERB} category which was most often confused with the \textit{NOUN} category. This behavior contradicts the naive linguistic assumptions. However, it can be justified by the fact that in agglutinating languages, verbs and nouns often have similar sets of affixes (e. g., possessive suffixes of nouns versus verbal personal suffixes).

Interestingly, both on the Evenki and Selkup dataset the quality of POS detection was comparable or even lower than the quality of morphological features detection. It contrasts the traditional ratio between the quality of recovery for POS tagging and morphological features: usually it is much easier to recover correct parts of speech than to restore all features. For example, during CoNLL2018 evaluation campaign \cite{zeman2018conll}, best average POS accuracy was $90.9\%$, while the accuracy of features was only $87.59\%$. Naturally, for Russian it is much easier to detect whether a word is a noun or a verb than to discriminate between, for example, accusative and nominative cases. Probably, this unusual performance can be explained by the abundancy of informal speech in the dataset, which is relatively ``unconnected'' in comparison to more formal sources of most UD treebanks. This implies that basic contextual clues (such as word order) prove too weak to predict part-of-speech labels. The corpora contain phrases with slips, repetitions, discourse markers, e. g.:
\textit{Wot amakalwi Ekondaduk bal= ekun kergentin, kergentin Ekondaduk} (So my grandfathers from Ekonda SLIP well, their family, their family from Ekonda)

\subsection{Morpheme segmentation}

The primary causes of the morpheme segmentation errors were the following:
\begin{enumerate}
    \item Non-standard and borrowed lemmata: as with the morphological analysis task, loanwords cause problems, with the systems splitting them incorrectly.
    On the other hand, loanwords with native suffixes such as \textit{telogrejka-t} 'coat(<Russian)-INSTR'
    \item Suffix combinations versus complex suffixes:
    interestingly, the \textbf{MSU-DeepPavlov} system sometimes splits a complex suffix into parts, e. g. \textit{d'eli} versus \textit{d'e-li}. Actually, the etymology of the suffix supports the claim that historically it could have been made of these basic parts. However, in the synchronous view, we cannot split the suffix.
\end{enumerate}
\subsection{Word form generation}
In the word form generation task, most errors were due to the vowel harmony and consonant alternation phenomena. Vowel harmony means that there are different forms of the same affix depending on the vowels in the stem. E. g., \textit{d'aja-} requires \textit{a} in some affixes. However, the system suggests \textit{e}, which is not correct.
It is worth noting that these phenomena are hard to grasp even in detailed grammatical descriptions. There is much variation in dialectal data. Sometimes the training data and gold standard data contradict the "normal" rules, so the results are not surprising. 

However, some errors cannot be justified by the data complexity as the resulting letter clusters are highly improbable and cannot be found in the training data.

\section{Conclusion}
In this paper, we present the results of the First Shared Task on morphology for low-resource languages. As a result of the shared task, several datasets in the CONLL format were prepared, for the first time for the languages in question. The participating teams created new morphological analysis tools for the languages which lack modern NLP technology tools. The comparison of results showed the vitality of modern neural approach when applied to low-resource datasets collected by field linguists. We also explored the limitations of the systems, which can help improve them.

\section{Acknowledgements}
The work of Elena Klyachko was partially supported by a grant of the Russian Science Foundation, Project 17-18-01649. 

The work of Natalia Krizhanovskaya and Andrew Krizhanovsky was supported by a grant of the Russian Foundation for Basic Research, Project 18-012-00117.

We are grateful to Svetlana Toldova, Artyom Sorokin and Karina Mishchenkova for their advice, and their help with evaluation scripts and datasets.

\medskip
\FloatBarrier
\begin{bibliography}{dialogue}
\end{bibliography}
 
\end{document}